\pdfoutput=1 

\documentclass[journal]{IEEEtran}
\ifCLASSINFOpdf
\else
\fi
\ifCLASSINFOpdf 
\usepackage[pdftex]{graphicx}
\else
\usepackage[dvips]{graphicx}
\fi

\usepackage{subfigure}
\usepackage{amsmath}
\usepackage{algorithmic}
\usepackage{array}

\usepackage{multirow}
\usepackage{algorithm,algorithmic}

\usepackage{url}
\begin{document}
\title{Image Based Appraisal of Real Estate Properties}

\author{Quanzeng You,
	Ran~Pang,
	Liangliang~Cao,
	and~Jiebo~Luo,~\IEEEmembership{Fellow,~IEEE}% <-this % stops a space
	\thanks{Copyright (c) 2013 IEEE. Personal use of this material is permitted. However, permission to use this material for any other purposes must be obtained from the IEEE by sending a request to pubs-permissions@ieee.org.}
	\thanks{Manuscript received March 28, 2016; accepted February 10, 2017.}
	\thanks{Q. You and J. Luo are with the Department
		of Computer Science, University of Rochester, Rochester, NY 14623 USA. E-mail:  \{qyou, jluo\}@cs.rochester.edu.}
	\thanks{R. Pang is with PayPaI. E-mail: pangrr89@gmail.com}
	\thanks{L. Cao is with Electrical Engineering \& Computer Sciences, Columbia University and customerserviceAI. E-mail: liangliang.cao@gmail.com}
}

\maketitle
\IEEEtitleabstractindextext{%
	\begin{abstract}
		Real estate appraisal, which is the process of estimating the price for real estate properties, is crucial for both buys and sellers as the basis for negotiation and transaction. Traditionally, the repeat sales model has been widely adopted to estimate real estate price. However, it depends the design and calculation of a complex economic related index, which is challenging to estimate accurately. Today, real estate brokers provide easy access to detailed online information on real estate properties to their clients. We are interested in estimating the real estate price from these large amounts of easily accessed data. In particular, we analyze the prediction power of online house pictures, which is one of the key factors for online users to make a potential visiting decision. The development of robust computer vision algorithms makes the analysis of visual content possible. In this work, we employ a Recurrent Neural Network (RNN) to predict real estate price using the state-of-the-art visual features. The experimental results indicate that our model outperforms several of other state-of-the-art baseline algorithms in terms of both mean absolute error (MAE) and mean absolute percentage error (MAPE).
	\end{abstract}
	
	}

% \fi
%\IEEEpeerreviewmaketitle

\section{Introduction}
\IEEEPARstart{R}eal estate appraisal, which is the process of estimating the price for real estate properties, is crucial for both buys and sellers as the basis for negotiation and transaction. Real estate plays a vital role in all aspects of our contemporary society. In a report published by the European Public Real Estate Association (EPRA \url{http://alturl.com/7snxx}), it was shown that \textit{real estate in all its forms accounts for nearly 20\% of the economic activity}. Therefore, accurate prediction of real estate prices or the trends of real estate prices help governments and companies make informed decisions. On the other hand, for most of the working class, housing has been one of the largest expenses. A right decision on a house, which heavily depends on their judgement on the value of the property, can possibly help them save money or even make profits from their investment in their homes. From this perspective, real estate appraisal is also closely related to people's lives.
\begin{figure}[!t]
	\centering
	\includegraphics[width=.45\textwidth]{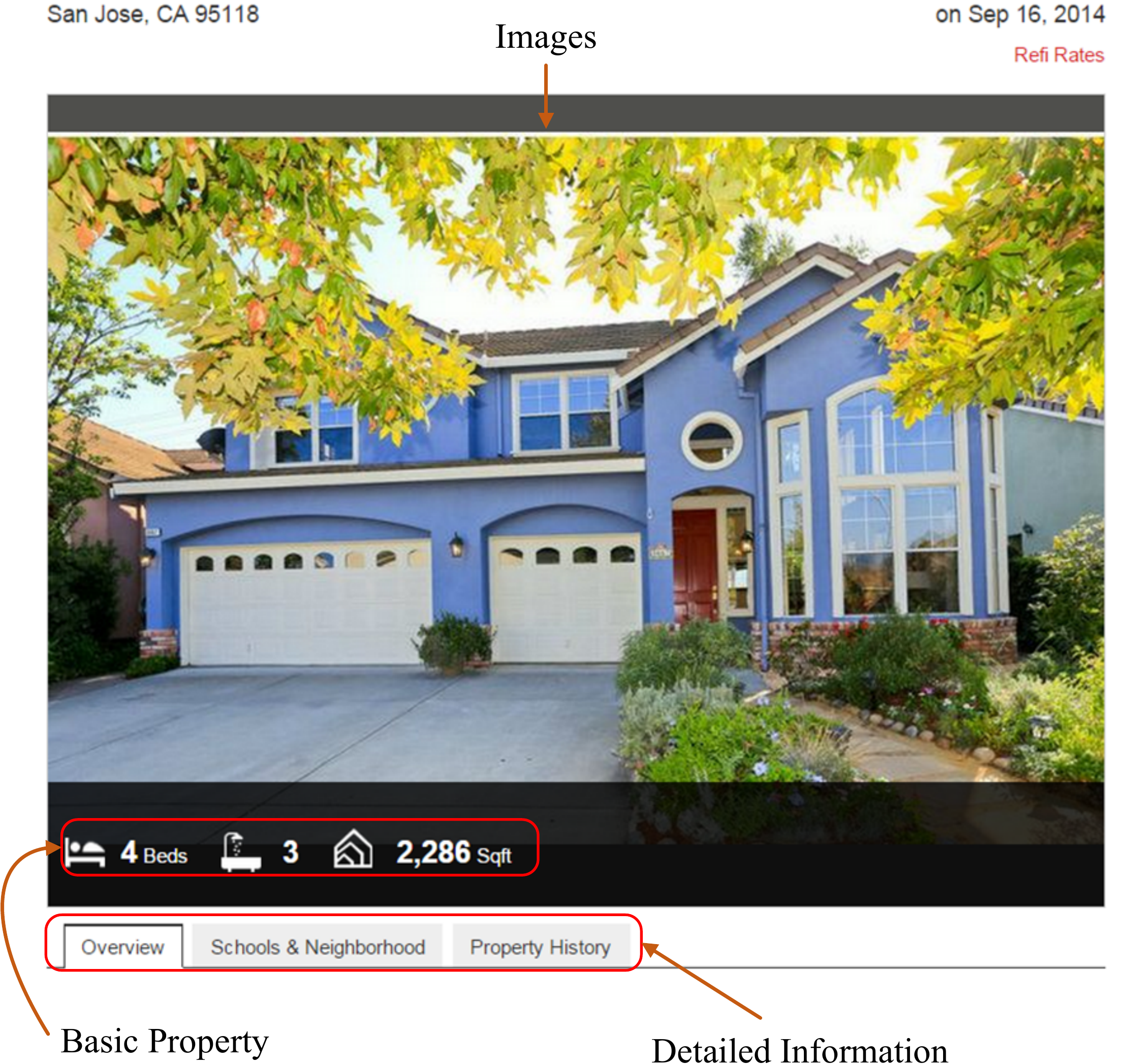}
	\caption{Example of homes for sale from Realtor.}
	\label{fig:house}
\end{figure}

Current research from both estate industry and academia has reached the conclusion that real estate value is closely related to property infrastructure~\cite{fu2014exploiting}, traffic~\cite{wardrip2011public}, online user reviews~\cite{Fu_Ge_Zheng_Yao_Liu_Xiong_Yuan_1970} and so on.
Generally speaking, there are several different types of appraisal values. In particular, we are interested in the \textit{market value}, which refers to the trade price \textit{in a competitive Walrasian auction setting}~\cite{beja1980dynamic}. Today, people are likely to trade through real estate brokers, who provide easy access online websites for browsing real estate property in an interactive and convenient way. \figurename~\ref{fig:house} shows an example of house listing from Realtor (\url{http://www.realtor.com/}), which is the largest real estate broker in North America. From the figure, we see that a typical piece of listing on a real estate property will introduce the infrastructure data in text for the house along with some pictures of the house. Typically, a buyer will look at those pictures to obtain a general idea of the overall property in a selected area before making his next move.

Traditionally, both real estate industry professionals and researchers have relied on a number of factors, such as economic index, house age, history trade and neighborhood environment~\cite{newurl} and so on to estimate the price. Indeed, these factors have been proved to be related to the house price, which is quite difficult to estimate and sensitive to many different human activities. Therefore, researchers have devoted much effort in building a robust house price index~\cite{bailey1963regression,meese1991nonparametric,sheppard1999hedonic,nagaraja2011autoregressive}. In addition, quantitative features including \textit{Area, Year, Storeys, Rooms} and \textit{Centre}~\cite{lasota2011empirical,kempa2011investigation} are also employed to build neural network models for estimating house prices. However, pictures, which is probably the most important factor on a buyer's initial decision making process~\cite{di2014picture}, have been ignored in this process. This is partially due to the fact that visual content is very difficult to interpret or quantify by computers compared with human beings.

%However, recent success of deep learning, in particular Convolutional Neural Networks, has introduced a wide range of computer vision related applications.
\emph{A picture is worth a thousand words}. One advantage with images and videos is that they act like universal languages. People with different backgrounds can easily understand the main content of an image or video. In the real estate industry, pictures can easily tell people exactly how the house looks like, which is impossible to be described in many ways using language. For the given house pictures, people can easily have an overall feeling of the house, \textit{e.g.} what is the overall construction style, how the neighboring environment looks like. These high-level attributes are difficult to be quantitatively described. On the other hand, today's computational infrastructure is also much cheaper and more powerful to make the analysis of computationally intensive visual content analysis feasible. Indeed, there are existing works on focusing the analysis of visual content for tasks such as prediction~\cite{jin2010wisdom,7293668}, and online user profiling~\cite{You2015}. Due to the recently developed deep learning, computers have become smart enough to interpret visual content in a way similar to human beings.

Recently, deep learning has enabled robust and accurate feature learning, which in turn produces the state-of-the-art performance on many computer vision related tasks, \textit{e.g.}~digit recognition~\cite{lecun1989backpropagation,hinton2006fast}, image classification~\cite{cirecsan2011flexible,krizhevsky2012imagenet}, aesthetics estimation~\cite{lu2014rapid} and scene recognition~\cite{zhou2014learning}. These systems suggest that deep learning is very effective in learning robust features in a supervised or unsupervised fashion. Even though deep neural networks may be trapped in local optima~\cite{hinton2010practical,bengio2012practical}, using different optimization techniques, one can achieve the state-of-the-art performance on many challenging tasks mentioned above.

Inspired by the recent successes of deep learning, in this work we are interested in solving the challenging real estate appraisal problem using deep visual features. In particular, for images related tasks, Convolutional Neural Network (CNN) are widely used due to the usage of convolutional layers. It takes into consideration the locations and neighbors of image pixels, which are important to capture useful features for visual tasks. Convolutional Neural Networks~\cite{lecun1998gradient,cirecsan2011flexible,krizhevsky2012imagenet} have been proved very powerful in solving computer vision related tasks.

We intend to employ the pictures for the task of real estate price estimation. We want to know whether visual features, which is a reflection of a real estate property, can help estimate the real estate price. Intuitively, if visual features can characterize a property in a way similar to human beings, we should be able to quantify the house features using those visual responses. Meanwhile, real estate properties are closely related to the neighborhood. In this work, we develop algorithms which only rely on 1) the neighbor information and 2) the attributes from pictures to estimate real estate property price.

To preserve the local relation among properties we employ a novel approach, which employs random walks to generate house sequences. In building the random walk graph, only the locations of houses are utilized. In this way, the problem of real estate appraisal has been transformed into a sequence learning problem. Recurrent Neural Network (RNN) is particularly designed to solve sequence related problems. Recently, RNNs have been successfully applied to challenging tasks including machine translation~\cite{bahdanau2014neural}, image captioning~\cite{vinyals2014show}, and speech recognition~\cite{graves2013speech}. Inspired by the success of RNN, we deploy RNN to learn regression models on the transformed problem.

The main contributions of our work are as follows:
\begin{itemize}
	\item{To the best of our knowledge, we are the first to quantify the impact of visual content on real estate price estimation. We attribute the possibility of our work to the newly designed computer vision algorithms, in particular Convolutional Neural Networks (CNNs).}
	\item{We employ random walks to generate house sequences according to the locations of each house. In this way, we are able to transform the problem into a novel sequence prediction problem, which is able to preserve the relation among houses.}
	\item{We employ the novel Recurrent Neural Networks (RNNs) to predict real estate properties and achieve accurate results.}
\end{itemize}
\section{Related Work}
\label{sec:related}
Real estate appraisal has been studied by both real estate industrial professionals and academia researchers. Earlier work focused on building price indexes for real properties. The seminal work in~\cite{bailey1963regression} built price index according to the repeat prices of the same property at different times. They employed regression analysis to build the price index, which shows good performances. Another widely used regression model, Hedonic regression, is developed on the assumption that the characteristics of a house can predict its price~\cite{meese1991nonparametric,sheppard1999hedonic}. However, it is argued that the Hedonic regression model requires more assumptions in terms of explaining its target~\cite{wang1997estimating}. They also mentioned that for repeat sales model, the main problem is lack of data, which may lead to failure of the model. Recent work in~\cite{nagaraja2011autoregressive} employed locations and sale price series to build an autoregressive component. Their model is able to use both single sale homes and repeat sales homes, which can offer a more robust sale price index.

More studies are conducted on employing feed forward neural networks for real estate appraisal~\cite{worzala1995exploration,rossini1998improving,kershaw1999using,nghiep2001predicting}. However, their results suggest that neural network models are unstable even using the same package with different run times~\cite{worzala1995exploration}. The performance of neural networks are closely related to the features and data size~\cite{nghiep2001predicting}. Recently, Kontrimas and Verikas~\cite{kontrimas2011mass} empirically studied several different models on selected $12$ dimensional features, \textit{e.g.} type of the house, size, and construction year. Their results show that linear regression outperforms neural network on their selected $100$ houses.

More recent studies in~\cite{fu2014exploiting} propose a ranking objective, which takes geographical individual, peer and zone dependencies into consideration. Their method is able to use various estate related data, which helps improve their ranking results based on properties' investment values. Furthermore, the work in~\cite{Fu_Ge_Zheng_Yao_Liu_Xiong_Yuan_1970} studied online user's reviews and mobile users' moving behaviors on the problem of real estate ranking. Their proposed sparsity regularized learning model demonstrated competitive performance.

In contrast, we are trying to solve this problem using the attributes reflected in the visual appearances of houses. In particular, our model does not use the meta data of a house (\textit{e.g.} size, number of rooms, and construction year). We intend to utilize the location information in a novel way such that our model is able to use the state-of-the-art deep learning for feature extraction (Convolutional Neural Network) and model learning (Recurrent Neural Network).

\section{Recurrent Neural Network for Real Estate Price Estimation}
In this section, we present the main components of our framework. We describe how to transform the problem into a problem that can be solved by the Recurrent Neural Network. The architecture of our model is also presented.
\subsection{Random Walks}
\label{sec:random}
One main feature of real estate properties is its location. In particular, for houses in the same neighborhood, they tend to have similar \textit{extrinsic} features including traffic, schools and so on. We build an undirected graph $G$ for all the houses collected, where each node $v_i$ represent the $i$-th house in our data set. The similarity $s_{ij}$ between house $h_i$ and house $h_j$ is defined using the \textit{Gaussian} kernel function, which is a widely used similarity measure\footnote{\url{http://en.wikipedia.org/wiki/Radial_basis_function_kernel}}:
\begin{equation}
s_{ij} = \exp\left(\frac{dist(h_i, h_j)}{2\sigma^2}\right),
\label{eqn:gaussian}
\end{equation}
where $dist(h_i,h_j)$ is the geodesic distance between house $h_i$ and $h_j$. $\sigma$ is the hyper-parameter, which controls the similarity decaying velocity with the increase of distance. In all of our experiments, we set $\sigma$ to 0.5 miles so that houses within the $1.5$ (within $3\sigma$) miles will have a relatively larger similarity. The $\epsilon$-neighborhood graph~\cite{von2007tutorial} is employed to build $G$ in our implementation. We assign the weight of each edge $e_{ij}$ as the similarity $s_{ij}$ between house $h_i$ and the house $h_j$.

Given this graph $G$, we can then employ random walks to generate sequences. In particular, every time, we randomly choose one node $v_i$ as the root node, then we proportionally jump to its neighboring nodes $v_j$ according to the weights between $v_i$ and its neighbors. The probability of jumping to node $v_j$ is defined as
\begin{equation}
p_j = \frac{e_{ji}}{\sum_{k \in N(i)} e_{ki}},
\label{eqn:random}
\end{equation}
where $N(i)$ is the set of neighbor nodes of $v_i$. We continue to employ this process until we generate the desired length of sequence. The employment of random walks is mainly motivated by the recent proposed DeepWalk~\cite{perozzi2014deepwalk} to learn feature representations for graph nodes. It has been shown that random walks can capture the local structure of the graphs. In this way, we can keep the local location structure of houses and build sequences for houses in the graph. Algorithm~\ref{alg:random} summarizes the detailed steps for generating sequences from a similarity graph.

We have generated sequences by employing random walks. In each sequence, we have a number of houses, which is related in terms of their locations. Since we build the graph on top of house locations, the houses within the same sequence are highly possible to be close to each other. In other words, the prices of houses in the same sequence are related to each other. We can employ this context for estimating real estate property price, which can be solved by recurrent neural network discussed in following sections.
\subsection{Recurrent Neural Network}
\begin{figure}[!htbp]
	\centering
	\includegraphics[width=0.45\textwidth]{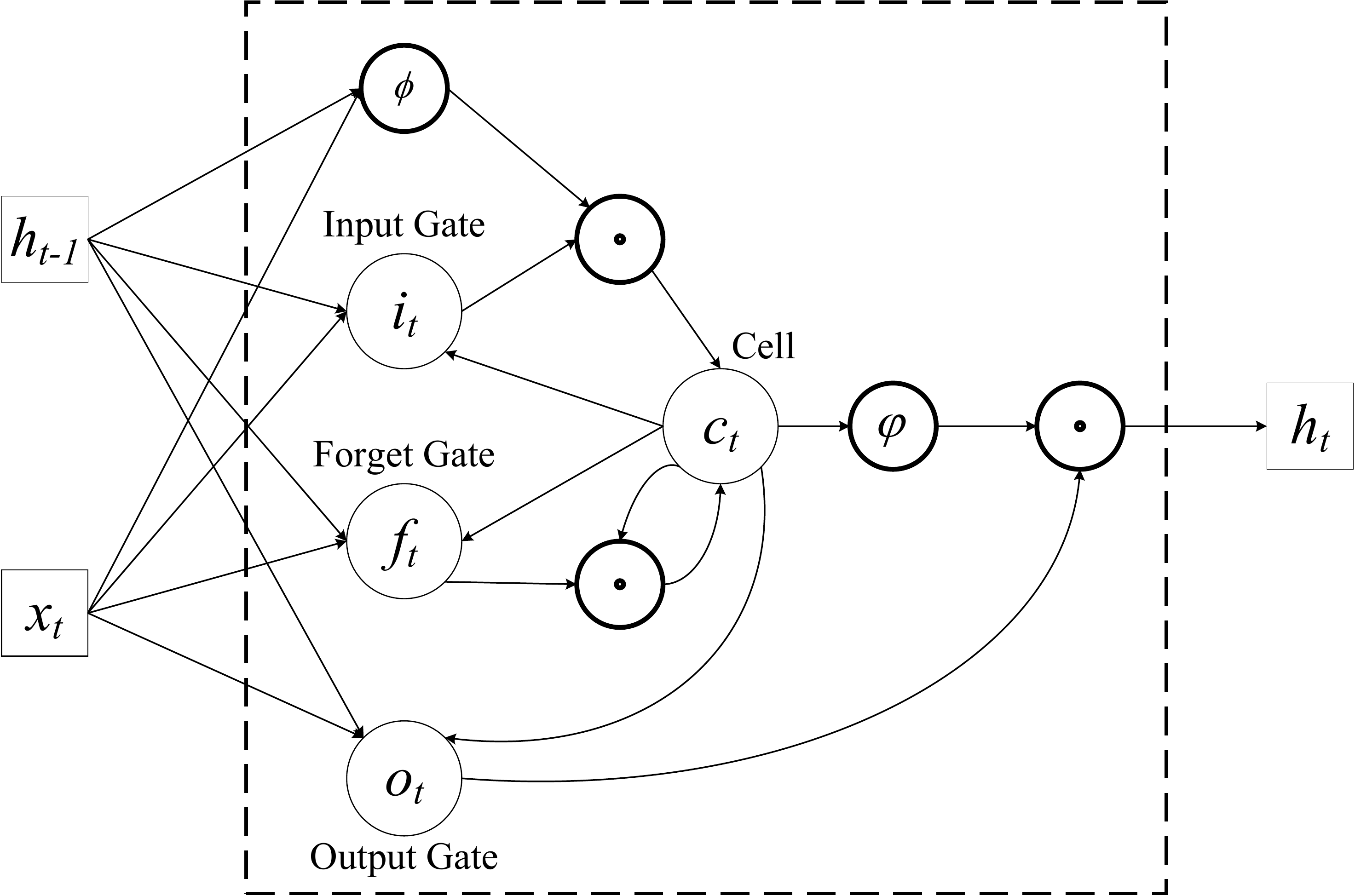}
	\caption{An illustration of a single Long Short-Term Memory (LSTM) Cell.}
	\label{fig:lstm}
\end{figure}

With a Recurrent Neural Network (RNN), we are trying to predict the output sequence $\{y_1, y_2, \dots, y_T\}$ given the input sequence $\{x_1,x_2, \dots, x_T\}$. Between the input layer and the output layer, there is a hidden layer, which is usually estimated as in Eq.(\ref{eqn:hidden}).
\begin{equation}
h_t = \Delta( W^i_h h_{t-1} + W_x x_t + b_h )
\label{eqn:hidden}
\end{equation}
$\Delta$ represents some selected activation function or other complex architecture employed to process the input $x_t$ and $h_t$. One of the most widely deployed architectures is Long Short-Term Memory (LSTM) cell~\cite{gers2001long}, which can overcome the \textit{vanishing} and \textit{exploding} gradient problem~\cite{pascanu2013difficulty} when training RNN with gradient descent. \figurename~\ref{fig:lstm} shows the details of a single Long Short-Term Memory (LSTM) block~\cite{gers2003learning}. Each LSTM cell contains an input gate, an output gate and an forget gate, which is also called a \textit{memory cell} in that it is able to \textit{remember} the error in the error propagation stage~\cite{hochreiter1997long}. In this way, LSTM is capable of modeling long-range dependencies than conventional RNNs.

\begin{figure*}[!htb]
	\centering
	\includegraphics[width=0.75\textwidth]{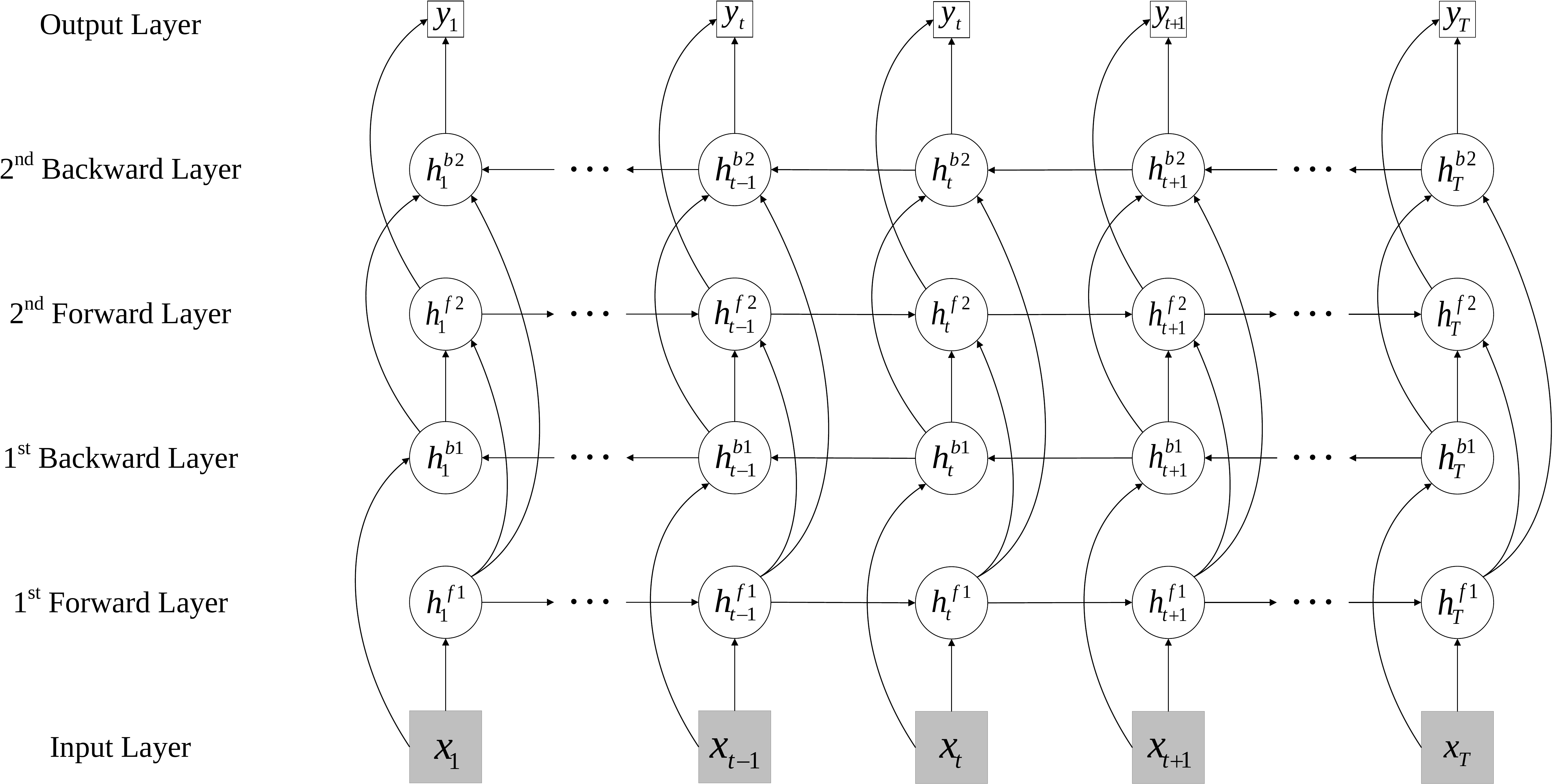}
	\caption{The Multi-layer Bidirectional Recurrent Neural Network (BRNN) architecture for real estate price estimation. There are two bidirectional recurrent layers in this architecture. For real estate price estimation, the price of each house is related to all houses in the same \emph{sequence}, which is the main motivation to employ bidirectional recurrent layers.}
	\label{fig:brnn}
\end{figure*}
For completeness, we give the detailed calculation of $h_t$ given input $x_t$ and $h_{t-1}$ in the following equations. Let $W_{.}^i$, $W_{.}^f$, $W_{.}^o$ represent the parameters related to input, forget and output gate respectively. $\odot$ denotes the element-wise multiplication between two vectors. $\phi$ and $\psi$ are some selected activation functions and $\sigma$ is the fixed logistic sigmoid function. Following~\cite{gers2003learning,graves2013speech,graves2013hybrid}, we employ $\tanh$ for both $\phi$ in Eq.(\ref{eqn:ct}) and $\psi$ in Eq.(\ref{eqn:ht}).
\begin{eqnarray}
&&i_t= \sigma(W_{x}^i x_t + W_{h}^i h_{t-1} + W_{c}^i c_{t-1} + b_i)\\
&&f_t = \sigma ( W_{x}^f x_t + W_{h}^f h_{t-1} + W_{c}^f c_{t-1} + b_f) \\
&&c_t = f_t \odot c_{t-1} + i _t \odot \phi(W_{x}^c x_t + W^c_h h_{t-1} + b_c) \label{eqn:ct}\\
&&o_t = \sigma(W_x^o x_t + W_h^o h_{t-1} + W_c^o c_t + b_o) \\
&&h_t = o_t \odot \psi(c_t) \label{eqn:ht}
\end{eqnarray}

\subsection{Multi-layer Bidirectional LSTM}
In previous sections, we have discussed the generation of sequences as well as Recurrent Neural Network. Recall that we have built an undirected graph in generating the sequences, which indicates that the price of one house is related to all the houses in the same sequence including those in the later part. Bidirectional Recurrent Neural Network (BRNN)~\cite{schuster1997bidirectional} has been proposed to enable the usage of both earlier and future contexts. In bidirectional recurrent neural network, there is an additional backward hidden layer iterating from the last of the sequence to the first. The output layer is calculated by employing both forward and backward hidden layer.

Bidirectional-LSTM (B-LSTM) is a particular type of BRNN, where each hidden node is calculated by the long short-term memory as shown in \figurename~\ref{fig:lstm}. Graves~\textit{et al.}~\cite{graves2013hybrid} have employed Bidirectional-LSTM for speech recognition. \figurename~\ref{fig:brnn} shows the architecture of the bidirectional recurrent neural network. We have two Bidirectional-LSTM layers. During the forward pass of the network, we calculate the response of both the forward and the backward hidden layers in the 1st-LSTM and 2nd-LSTM layer respectively. Next, the output (in our problem, the output is the price of each house) of each house is calculated using the output of the 2nd-LSTM layer as input to the output layer.

\begin{algorithm}[!htp]
	\caption{RandomWalks}
	\label{alg:random}
	\begin{algorithmic}[1]
		\REQUIRE $H=\{h_1,h_2,\dots,h_n\}$ geo-coordinates of $n$ houses\\
		\quad \ \ $\sigma$ hyper-parameter for Gaussian Kernel \\
		\quad \ \ $t$ threshold for distance \\
		\quad \ \ $M$ total number of desired sequences \\
		\STATE Calculate the Vincenty distance between any pair of houses
		\STATE Calculate the similarity between houses according to the Gaussian kernel function (see Eq.(\ref{eqn:gaussian})).
		\REPEAT
		\STATE Initialize $s_c = \{\}$
		\STATE Randomly pick one node $h_i$ and add $h_i$ to $s_c$
		\STATE set $h_c = h_i$
		\WHILE{size$(s_c) < L$}
		\STATE Pick $h_c$'s neighbor node $h_j$ with probability $p_j$ defined in Eq.(\ref{eqn:random})
		\STATE add $h_j$ to $s_c$
		\STATE set $h_c = h_j$
		\ENDWHILE
		add $s_c$ to $S$
		\UNTIL{size $(S) = M$}
		\RETURN The set of sequence $S$
	\end{algorithmic}
\end{algorithm}

The objective function for training the Multi-Layer Bidirectional LSTM is defined as follows:
\begin{eqnarray}
L = \frac{1}{N} \sum_{n=1}^{N} \sum_j \parallel \hat{y}_{ij} - y_{ij}\parallel^2% + \lambda |W| +  \gamma \parallel W \parallel_2,
\label{eqn:loss}
\end{eqnarray}
where $W$ is the the set of all the weights between different layers.%, and $|\cdot|$ and $||\cdot||_2$ are the $l1$-norm and $l2$-norm of each weight matrix individually.
$y_{ij}$ is the actual trade price for the $j$-th house in the generated $i$-th sequence and $\hat{y}_{ij}$ is the corresponding estimated price for this house.

When training our Multi-Layer B-LSTM model, we employ the RMSProp~\cite{rmsprop} optimizer, which is an adaptive method for automatically adjust the learning rates. In particular, it normalizes the gradients by the average of its recent magnitude.
%The following equations give the update of parameter $W$ in one iteration step.
%\begin{eqnarray}
%&&\xi_t = (1 - \lambda) \xi_{t-1} + \lambda \parallel \frac{\partial L }{\partial W}\parallel^2 \\
%&& \Delta W = \frac{\partial L}{\partial W} \\
%&& W_{t+1} = W_t - \alpha \Delta W \label{eqn:update}
%\end{eqnarray}

\begin{algorithm}[!htp]
	\caption{Training Multi-Layer B-LSTM}
	\label{alg:blstm}
	\begin{algorithmic}[1]
		\REQUIRE  $H=\{h_1,h_2,\dots,h_n\}$ geo-coordinates of $n$ houses\\
		\quad \ \ $X=\{x_1, x_2, \dots, x_n\}$ features of the $n$ house\\
		\quad \ \ $Y=\{y_1, y_2, \dots, y_n\}$ prices of the $n$ houses
		%$B \in \mathbf{R}^{p \times (n-p)}$: affinity matrix between sampled and\\ \quad \ remaining data points \\
		%\ENSURE $I=\{i_1,i_2,\dots,i_m\}$ the indices of sampled points \\
		\STATE $S$ = RandomWalks (see Algorithm~\ref{alg:random})
		\STATE Split $S$ into mini-batches
		\REPEAT
		\STATE Calculate the gradient of $L$ in Eq.(\ref{eqn:loss}) and update the parameters using RMSProp.
		\UNTIL Convergence
		\RETURN The learned model $M$
	\end{algorithmic}
\end{algorithm}
We conduct the back propagation in a mini-batch approach. Algorithm~\ref{alg:blstm} summarizes the main steps for our proposed algorithm.
% sub-alg: random walk.

\subsection{Prediction}
\begin{figure}[!htb]
	\centering
	\includegraphics[width=0.475\textwidth]{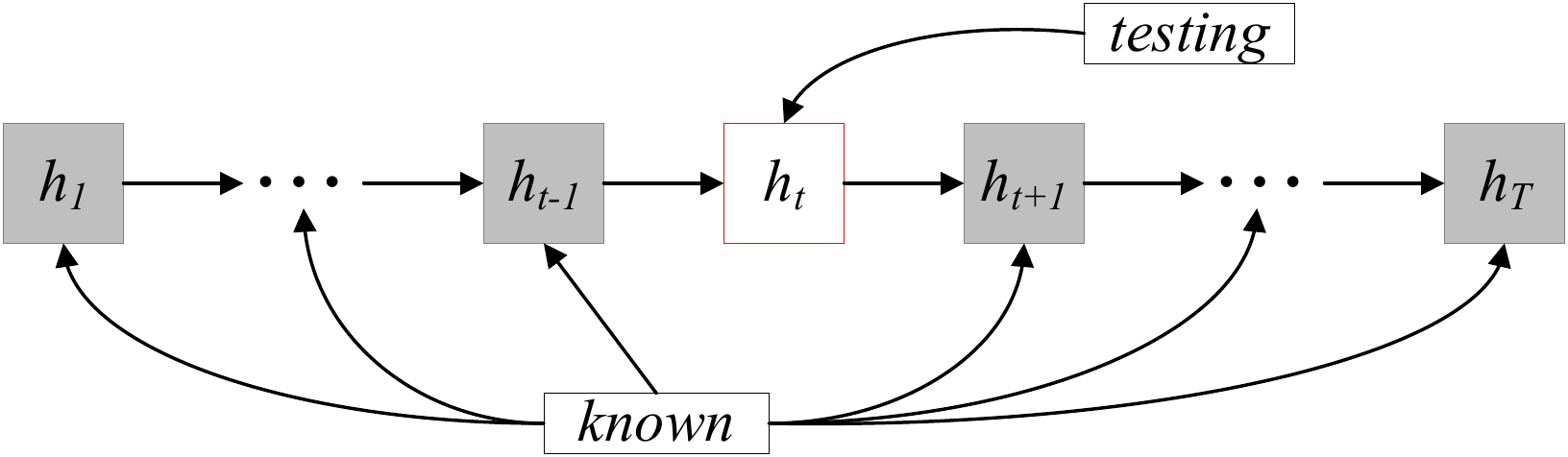}
	\caption{Testing sequence $h_1\rightarrow h_2 \rightarrow  \cdots \rightarrow  h_T$. In each testing sequence, there is one and only one testing node in that sequence. The remaining nodes are all come from training data.}
	\label{fig:seq}
\end{figure}

In the prediction stage, the first step is also generating sequence. For each testing house, we add it as a new node into our previously build similarity graph on the training data. Each testing house is a new node in the graph. Next, we add edges to the testing nodes and the training nodes. We use the same settings when adding edges to the new $\epsilon$-neighborhood graph. Given the new graph $G'$, we randomly generate sequences and keep those sequences that contain one and only one testing node. In this way, for each house, we are able to generate many different sequences that contain this house. \figurename~\ref{fig:seq} shows the idea. Each testing sequence only has one testing house. The remaining nodes in the sequence are the known training houses.

\paragraph{Average} The above strategy implies that we are able to build many different sequences for each testing house. To obtain the final prediction price for each testing house, one simple strategy is to average the prediction results from different sequences and report the average price as the final prediction price.

\section{Experimental Results}
\label{sec:exp}
In this section, we discuss how to collect data and evaluate the proposed framework as well as several state-of-the-art approaches. In this work, all the data are collected from Realtor~(\url{http://www.realtor.com/}), which is the largest realtor association in North America. We collect data from San Jose, CA, one of the most active cities in U.S., and Rochester, NY, one of the least active cities in U.S., over a period of one year.  In the next section, we will discuss the details on how to preprocess the data for further experiments.

\subsection{Data Preparation}
The data collected from Realtor contains description, school information and possible pictures about each real property as shown in \figurename~\ref{fig:house} show. We are particularly interested in employing the pictures of each house to conduct the price estimation. We filter out those houses without image in our data set. Since houses located in the same neighborhood seem to have similar price, the location is another important features in our data set. However, after an inspection of the data, we notice that some of the house price are abnormal. Thus, we preprocess the data by filtering out houses with extremely high or low price compared with their neighborhood.

\begin{figure*}
	\centering
	\subfigure[Rochester]{
		\centering
		\includegraphics[width=.45\textwidth]{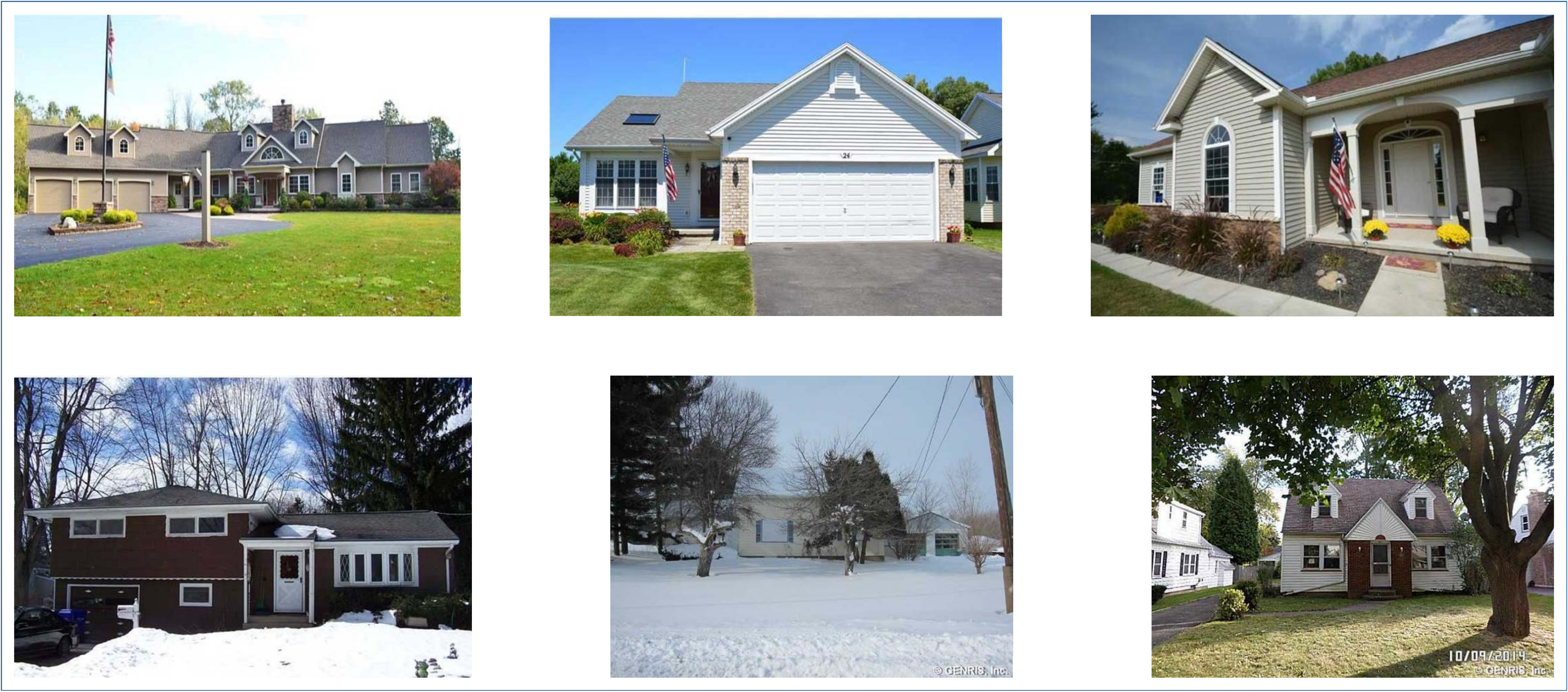}
	}
	\subfigure[San Jose]{
		\centering
		\includegraphics[width=.45\textwidth]{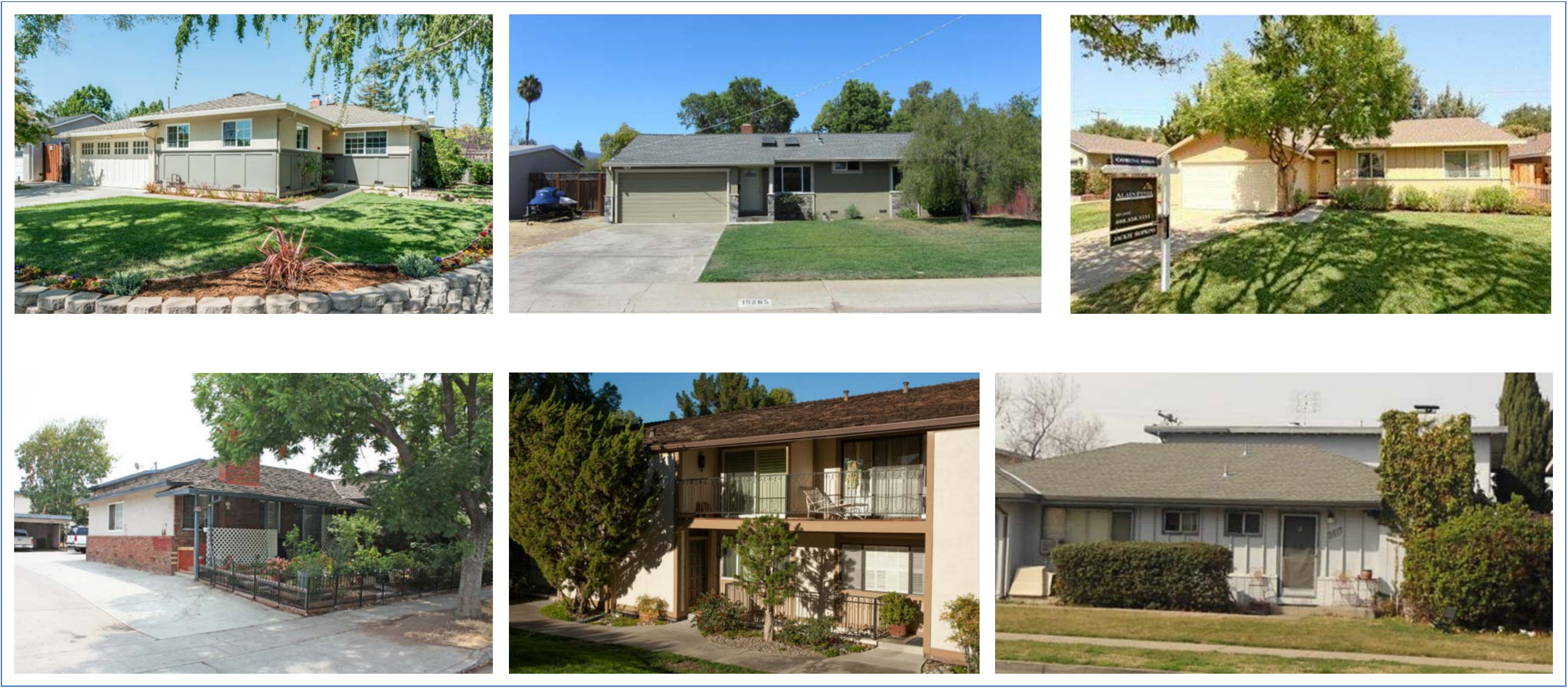}
	}
	\caption{Examples of house pictures of the two cities respectively. Top Row: houses whose prices (per Sqft) are above the average of their neighborhood. Bottom Row: houses whose prices (per Sqft) are below the average of their neighborhood.}
	\label{fig:houses}
\end{figure*}

\tablename~\ref{tab:data} shows the overall statistics of our dataset after filtering. Overall, the city of San Jose has more houses than
Rochester on the market (as expected for one of the hottest market in the country). The house prices in the two cities also have
significant differences. \figurename~\ref{fig:houses} shows some of the example
house pictures from the two cities, respectively. From these
pictures, we observe that houses whose prices are above average typically have larger yards and better curb appeal, and vice versa. The same can be observed among house interior pictures (examples not shown due to space).
%\tablename~\ref{tab:data} shows the overall statistics of our dataset after filtering. Overall, the city of San Jose has more houses than Rochester. The house prices in the two cities also have a significant difference. \figurename~\ref{fig:houses} shows some of the example house pictures from the two cities respectively. From these pictures, we observe that houses whose price is above average typically have large open area and the houses looks in good shape. In contrast, houses whose price is below average seem to be compact and sit in a relatively narrow neighborhood.
%%%%%%%%%%%%%%%%%%%%%%%%%%%%%%%%%%%%%%%%%%%%%%%%%%%%%%%%%%%%%
%%%%%%%%%%%%%%%%%%%%%%%%%%%%%%%%%%%%%%%%%%%%%%%%%%%%%%%%%%%%%
\begin{table}[!hbtp]
	\centering{
		\caption{The average price per Sqft and the standard deviation (std) of the price of the two studied cities.}
		\label{tab:data}
		\small
		\begin{tabular}{|l|l|l|l|} \hline
			City & \# of Houses & Avg Price & std of Price \\ \hline
			San Jose & 3064 & 454.2 &  132.1\\ \hline
			Rochester & 1500 & 76.4 & 21.2 \\ \hline
		\end{tabular}
	}
\end{table}

\begin{figure}[!htb]
	\centering
	\subfigure[San Jose, CA]{
		\centering
		\includegraphics[width=0.45\textwidth]{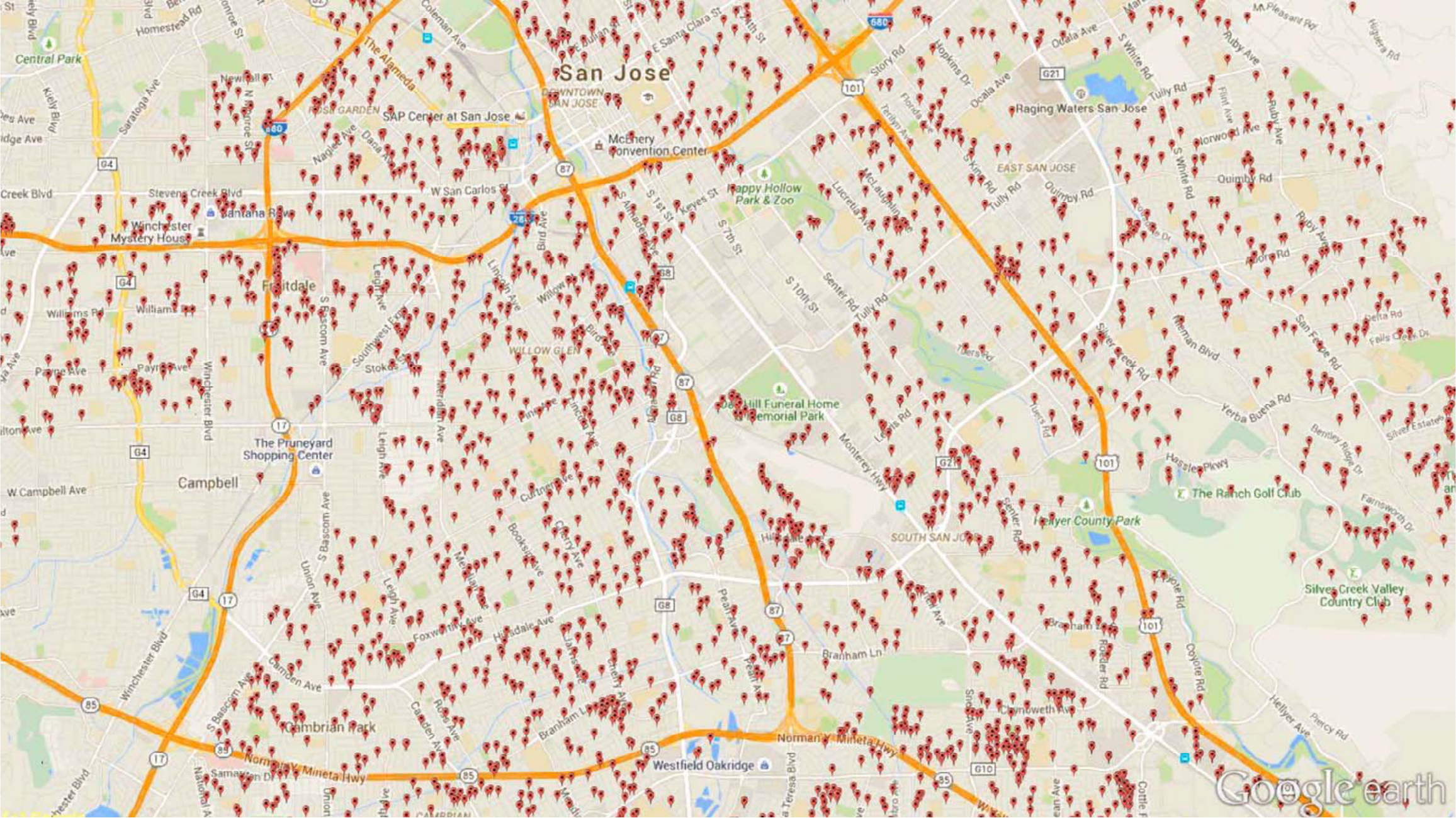}
		\label{fig:map:sj}
	}
	\subfigure[Rochester, NY]{
		\centering
		\includegraphics[width=0.45\textwidth]{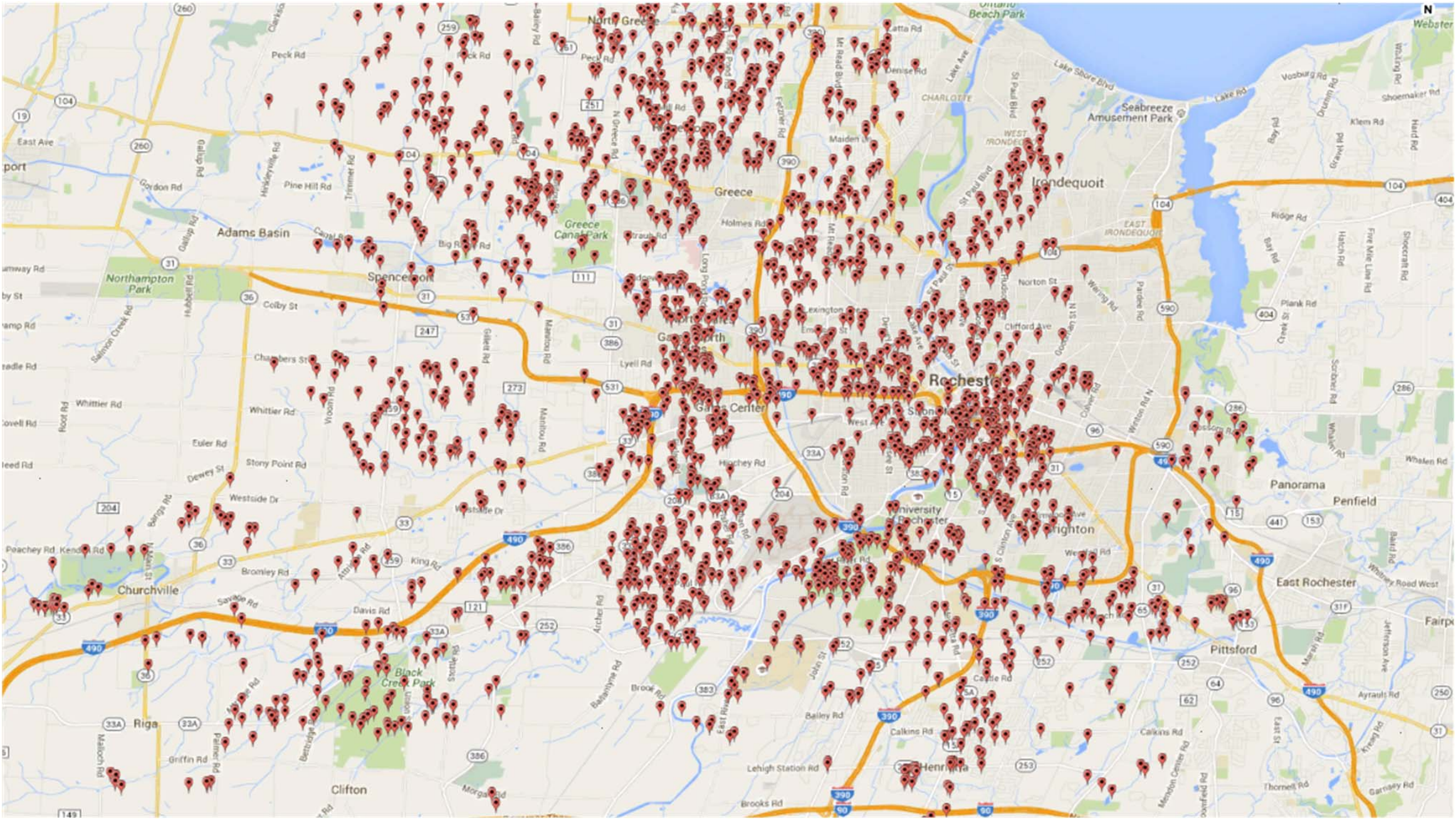}
		\label{fig:map:roc}
	}
	\caption{Distribution of the houses in our collected data for both San Jose and Rochester according to their geo-locations.}
	\label{fig:maps}
\end{figure}

Realtor does not provide the exact geo-location for each house. However, geo-location is important for us to build the $\epsilon$-neighborhood graph for random walks. We employ Microsoft Bing Map API~(\url{https://msdn.microsoft.com/en-us/library/ff701715.aspx}) to obtain the latitude and longitude for each house given its collected address. \figurename~\ref{fig:maps} shows some of the houses in our collected data from San Jose and Rochester using the returned geo-locations from Bing Map API.

\begin{figure}[!htb]
	\centering
	\includegraphics[width=0.4\textwidth]{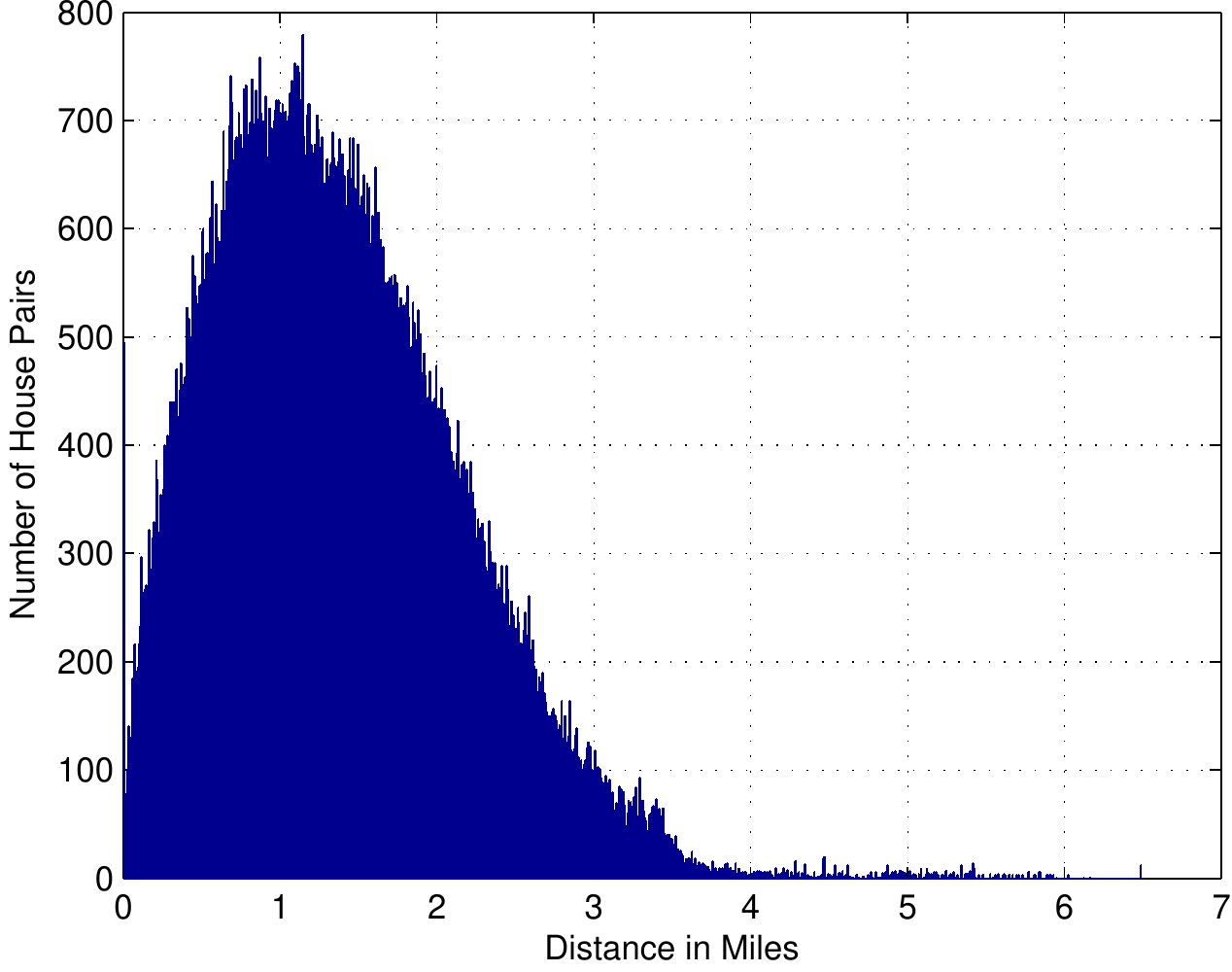}
	\caption{Distribution of distances between different pairs of houses.}
	\label{fig:dist}
\end{figure}

According to these coordinates, we are able to calculate the distance between any pair of houses. In particular, we employ Vincenty distance~(\url{https://en.wikipedia.org/wiki/Vincenty's_formulae}) to calculate the geodesic distances according to the coordinates. \figurename~\ref{fig:dist} shows distribution of the distance between any pair of houses in our data set. The distance is less than 4 miles for most randomly picked pair of houses. In building our $\epsilon$-neighborhood graph, we assign an edge between any pair of houses, which has a distance smaller than 5 miles ($\epsilon = 5$ miles).

\begin{table*}[!htbp]
	\centering{
		\caption{Prediction deviation of different models from the actual sale prices. Note that RNN-best is the upper-bound performance of the RNN based model proposed in this work.}
		\label{tab:reg}
		\small
		\begin{tabular}{*{9}{|l}|} \hline
			\multicolumn{1}{|c|}{\multirow{2}{*}{ City }} & \multicolumn{2}{c|}{LASSO} &
			\multicolumn{2}{c|}{DeepWalk} & \multicolumn{2}{c|}{RNN-best} &
			\multicolumn{2}{c|}{RNN-avg}\\
			\cline{2-9}
			& \multicolumn{1}{c|}{MAE} & \multicolumn{1}{c|}{MAPE} & \multicolumn{1}{c|}{MAE} & \multicolumn{1}{c|}{MAPE}& \multicolumn{1}{c|}{MAE} & \multicolumn{1}{c|}{MAPE} & \multicolumn{1}{c|}{MAE} & \multicolumn{1}{c|}{MAPE} \\ \hline
			San Jose & 70.79& 16.92\% & 68.05 & 16.12\% & \textbf{17.98} & \textbf{4.58\%} & \textbf{66.3} & \textbf{16.11\%} \\ \hline
			Rochester & 14.19 & 24.83\% & 13.68 & 23.28\% & \textbf{5.21} & \textbf{9.94\%} & \textbf{13.32} & \textbf{22.69\%} \\ \hline
		\end{tabular}
	}
\end{table*}

\subsection{Feature Extraction and Baseline Algorithms}
In our implementation, we experimented with GoogleNet model~\cite{Szegedy_2015_CVPR}, which is one of the state-of-the-art deep neural architectures. In particular, we use the response from the last $avg-pooling$ layer as the visual features for each image. In this way, we obtain a $1,024$ dimensional feature vector for each image. Each house may have several different pictures on different angles of the same property. We average features of all the images of the same house (also known as \textit{average-pooling})\footnote{We also tried max-pooling. However, the results are not as good as average-pooling. In the following experiments, we report the results using average-pooling.} to obtain the feature representation of the house.

We compare the proposed framework with the following algorithms.
\subsubsection{Regression Model (LASSO)}
Regression model has been employed to analyze real estate price index~\cite{bailey1963regression}. Recently, the results in Fu~\textit{et al.}~\cite{Fu_Ge_Zheng_Yao_Liu_Xiong_Yuan_1970} show that sparse regularization can obtain better performance in real estate ranking. Thus, we choose to use LASSO (\url{http://statweb.stanford.edu/~tibs/lasso.html}), which is a $l1$-constrained regression model, as one of our baseline algorithms.
\subsubsection{DeepWalk}
Deepwalk~\cite{perozzi2014deepwalk} is another way of employing random walks for unsupervised feature learning of graphs. The main approach is inspired by distributed word representation learning. In using DeepWalk, we also use $\epsilon$-neighborhood graph with the same settings with the graph we built for generating sequences for B-LSTM. The learned features are also fed into a LASSO model for learning the regression weights. Indeed, deepwalk can be thought as a simpler version of our algorithm, where only the graph structure are employed to learn features. Our framework can employ both the graph structure and other features, \textit{i.e.} visual attributes, for building regression model.
\subsection{Training a Multi-layer B-LSTM Model}
With the above mentioned similarity graph, we are able to generate sequences using random walks following the steps described in Algorithm~\ref{alg:random}. For each city, we randomly split the houses into training (80\%) and testing set (20\%). Next, we generate sequences using random walks on the training houses only to build our training sequences for Multi-layer B-LSTM.

For both cities, we build $200,000$ sequences for training, with a length of $10$. Similarly, we also generate testing sequences, where each sequence contain one and only one testing house (see \figurename~\ref{fig:seq}). On the average, we randomly generate $100$ sequences for each testing house. The B-LSTM model is trained with a batch size of $1024$. In our experimental settings, we set the size of the first hidden layer to be $400$ and the size of the second hidden layer to be $200$.

The evaluation metrics employed are mean absolute error (MAE) and mean absolute percentage error (MAPE). Both of them are popular measures for evaluating the accuracy of prediction models. Eq.(\ref{eqn:mae}) and Eq.(\ref{eqn:mape}) give the definitions for these two metrics, where $p_i$ is the predicted value and $t_i$ is the true value for the $i$-th instance.
\begin{align}
	\text{MAE}\ & = \frac{1}{N} \sum_{i=1}^{N} |t_i - p_i| \label{eqn:mae}\\
	\text{MAPE}\ & = \frac{1}{N} \sum_{i=1}^{N} \rvert  \frac{t_i - p_i}{t_i}\label{eqn:mape}\rvert
\end{align}

We use the same training and testing split to evaluate all the approaches. \tablename~\ref{tab:reg} shows the regression results for all the different approaches in the two selected cities. For each testing house, we generate about $100$ sequences. In \tablename~\ref{tab:reg}, we report both the best and the average price of the predicted price. For Rochester, the average standard deviation of the predicted prices over all the houses is $5.6$, which is $7.33\%$ of the average price in Rochester (see \tablename~{\ref{tab:data}}). Comparably, the average standard deviation for San Jose is $34.64$, which is $7.63\%$ of the average price in San Jose.  The \emph{best} is the price closest to the true price among all the available sequences for each house\footnote{This is the upper bound of the prediction results. We choose the closest price using the ground truth price as reference.}. Overall, our B-LSTM model outperforms other two baseline algorithms in both cities. All of the evaluation approaches perform better in San Jose than in Rochester in terms of MAPE. This is possible due to the availability of more training data in the city of San Jose. DeepWalk shows slightly better performance than LASSO, which suggests that location is relatively more important than the visual features in the realtor business. This is expected

\begin{figure}[!htb]
	\subfigure[MAE]{
		\centering
		\includegraphics[width=0.45\textwidth]{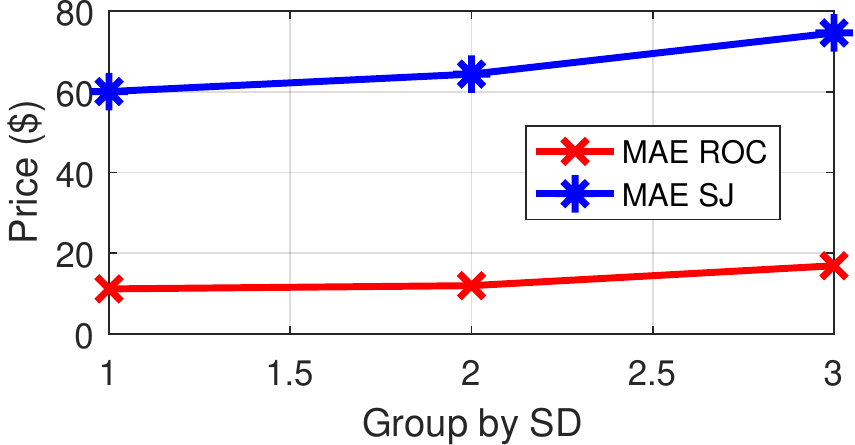}
		\label{fig:mae}
	}
	\subfigure[MAPE]{
		\centering
		\includegraphics[width=0.45\textwidth]{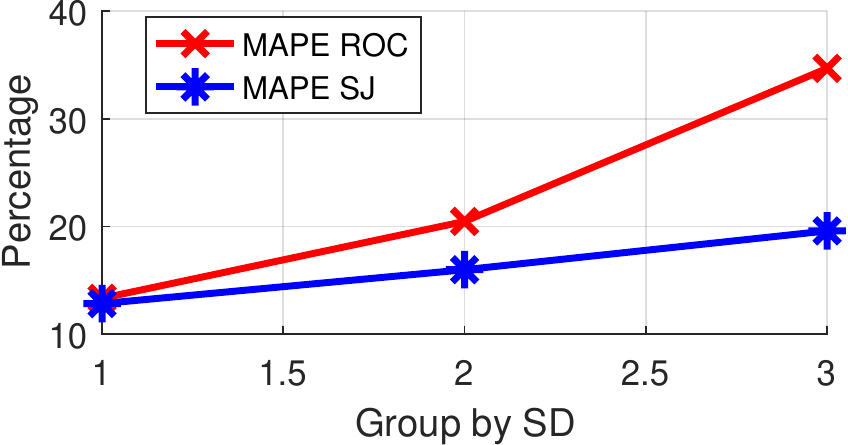}
		\label{fig:mape}
	}
	\caption{Performance of B-LSTM-avg in different groups. All the testing houses are grouped by the predicted standard deviation.}
	\label{fig:var}
\end{figure}

\subsection{Confidence Level}
For each testing house, the proposed model can give a group of predictions. We want to know whether or not the proposed model can distinguish the confidence level of its prediction. In particular, we group the testing houses evenly into three groups for each city. The first group has the smallest standard deviation of the prediction prices. The second group is the middle one and the last group is the one with the largest standard deviation.

\figurename~\ref{fig:var} shows the MAE and MAPE for the different groups. The results show that standard deviation can be viewed as a rough measure of the confidence level of the proposed model on the current testing house. Small standard deviation tends to indicate a high confidence of the model and overall it also suggests a smaller prediction error.

\section{Conclusion}
In this work, we propose a novel framework for real estate appraisal. In particular, the proposed framework is able to take both the location and the visual attributes into consideration. The evaluation of the proposed model on two selected cities suggests the effectiveness and flexibility of the model. Indeed, our work has also offered new approaches of applying deep neural networks on graph structured data. We hope our model can not only give insights on real estate appraisal, but also can inspire others on employing deep neural networks on graph structured data.

\ifCLASSOPTIONcaptionsoff
\newpage
\fi

\bibliographystyle{IEEEtran}
\bibliography{icwsm2016}
\end{document}